%% file: main.tex
\documentclass[conference]{IEEEtran}
\IEEEoverridecommandlockouts

\usepackage{geometry}

\usepackage{cite}
\usepackage{amsmath,amssymb,amsfonts}
\usepackage{algorithmic}
\usepackage{graphicx}
\usepackage{textcomp}
\usepackage[table,xcdraw]{xcolor}
\usepackage{listings}

\definecolor{codegreen}{rgb}{0,0.6,0}
\definecolor{codegray}{rgb}{0.5,0.5,0.5}
\definecolor{codepurple}{rgb}{0.58,0,0.82}
\definecolor{backcolour}{rgb}{0.95,0.95,0.92}
\usepackage{tcolorbox}

\lstdefinestyle{mystyle}{
    backgroundcolor=\color{backcolour},   
    commentstyle=\color{codegreen},
    keywordstyle=\color{magenta},
    numberstyle=\tiny\color{codegray},
    stringstyle=\color{codepurple},
    basicstyle=\ttfamily\footnotesize,
    breakatwhitespace=false,         
    breaklines=true,                 
    captionpos=b,                    
    keepspaces=true,                 
    numbers=left,                    
    numbersep=5pt,                  
    showspaces=false,                
    showstringspaces=false,
    showtabs=false,                  
    tabsize=2,
}

\usepackage{enumitem}

\def\BibTeX{{\rm B\kern-.05em{\sc i\kern-.025em b}\kern-.08em
    T\kern-.1667em\lower.7ex\hbox{E}\kern-.125emX}}

\makeatletter 
\newcommand{\linebreakand}{%
  \end{@IEEEauthorhalign}
  \hfill\mbox{}\par
  \mbox{}\hfill\begin{@IEEEauthorhalign}
}
\makeatother 

\definecolor{miro_light_blue}{RGB}{18, 205, 212}
\definecolor{miro_light_green}{RGB}{143, 209, 79}
\definecolor{miro_light_yellow}{RGB}{255, 246, 143}

\newcommand{\inlinegray}[1]{\textcolor[rgb]{0.3,0.3,0.3}{#1}}

\begin{document}
\newgeometry{top=1in,left=0.75in,right=0.75in,bottom=0.75in}

\title{Domain Knowledge Distillation from Large Language Model: An Empirical Study in the Autonomous Driving Domain\\
\thanks{All authors are with WMG, University of Warwick, Coventry, United Kingdom. *Corresponding author. 
}
}

\author{
\IEEEauthorblockN{Yun Tang*}
\IEEEauthorblockA{
yun.tang@warwick.ac.uk}
\and
\IEEEauthorblockN{Antonio A. Bruto da Costa}
\IEEEauthorblockA{
antonio.bruto-da-costa@warwick.ac.uk}
\and
\IEEEauthorblockN{Xizhe Zhang}
\IEEEauthorblockA{
Jason.Zhang@warwick.ac.uk}

\linebreakand
\IEEEauthorblockN{Irvine Patrick}
\IEEEauthorblockA{
patrick.irvine@warwick.ac.uk}
\and
\IEEEauthorblockN{Siddartha Khastgir}
\IEEEauthorblockA{
S.Khastgir.1@warwick.ac.uk}
\and
\IEEEauthorblockN{Paul Jennings}
\IEEEauthorblockA{
Paul.Jennings@warwick.ac.uk}
}

\maketitle

\begin{abstract}
\input{Abstract}
\end{abstract}


\input{Introduction}
\input{Methodology}
\input{Experiment}
\input{UI}
\input{Conclusion}

\bibliographystyle{ieeetr}
\bibliography{references} 

\end{document}

%% file: Abstract.tex
Engineering knowledge-based (or expert) systems require extensive manual effort and domain knowledge. As Large Language Models (LLMs) are trained using an enormous amount of cross-domain knowledge, it becomes possible to automate such engineering processes. This paper presents an empirical automation and semi-automation framework for domain knowledge distillation using prompt engineering and the LLM ChatGPT. We assess the framework empirically in the autonomous driving domain and present our key observations. In our implementation, we construct the domain knowledge ontology by “chatting” with ChatGPT. The key finding is that while fully automated domain ontology construction is possible, human supervision and early intervention typically improve efficiency and output quality as they lessen the effects of response randomness and the butterfly effect. We, therefore, also develop a web-based distillation assistant enabling supervision and flexible intervention at runtime. We hope our findings and tools could inspire future research toward revolutionizing the engineering of knowledge-based systems across application domains.

%% file: Introduction.tex
\section{Introduction}

Large language models (LLMs), such as GPT-3 \cite{brown2020language}, Codex \cite{chen2021evaluating}, and ChatGPT \cite{chatGPT} have made remarkable progress. Trained using an enormous amount of indiscriminate data from the entire internet, these LLMs embed knowledge from different domains, which are thus capable of answering questions, writing codes, drawing pictures, or translating languages across application areas \cite{vemprala2023chatgpt, pearce2022examining, schafer2023adaptive}. In this paper, we aim to investigate if and how the knowledge of a specific application domain, e,g., scenario-based testing of autonomous vehicles, can be extracted to facilitate subsequent tasks, e.g. automatic testing scenario generation.

Safety verification and validation (V\&V) of autonomous vehicles (AVs) are challenging due to the complexity of the AVs and their operating environment. Scenario-based testing of AVs \cite{zhong2021survey,tang2022survey} has been a new V\&V paradigm compared to distance-based approaches, where the performance of AVs is evaluated against the types of scenarios they pass instead of the countless miles they travel. 

Many scenario-generation methods have been proposed, e.g., \cite{tang2021route, tang2022automatic, zhou2023flyover, ding2022survey, tang2021collision, tang2021systematic}. However, those methods are mostly ``parameter samplers'' instead of ``scenario explorers'', meaning they are proposed to sample critical parameter values given a fixed list of scenario parameters toward their generation directions. Still, they cannot systematically explore different functional scenarios \cite{menzel2018scenarios}, e.g. different road networks, traffic actors, and their manoeuvres. Our recent work \cite{khastgir2021systems} applies \textit{Systems Theoretic Process Analysis} (STPA) to explore different scenario types at the functional scenario level; however, such a method requires ``domain knowledge'' and ``manual effort'' extensively and hence does not scale.

\begin{figure}
    \centering
    \includegraphics[width=\columnwidth]{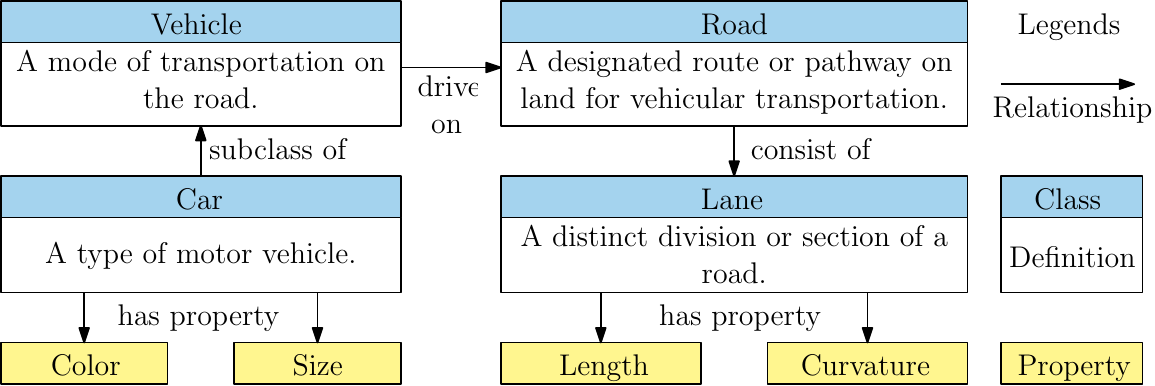}
    \caption{Visualization of an ontology example adapted from the OpenXOntology \cite{ASAM_OpenXOntology}, manually designed for road traffic domain.}
    \label{fig:asam_openxOntology_example}
\end{figure}

Recently, to eliminate the ``manual effort'', combinatorial sampling-based methods are proposed to systematically generate different scenarios given a form of domain knowledge, e.g., either \textit{Operational Design Domain} (ODD) \cite{weissensteiner2023operational} or \textit{Ontology} \cite{bagschik2018ontology}.
The scenario ontology (Figure~\ref{fig:asam_openxOntology_example}) is a form of domain knowledge aiming to encapsulate all the relevant physical entities, their relationships, as well as their associated events and activities, which thus has the potential to generate any scenario. 
However, no approaches have been proposed to automatically ``distil'' such ``domain knowledge'' in any form from scratch for subsequent automation tasks, such as scenario generation, until it becomes feasible with the recent progress in Artificial General Intelligence (AGI) \newgeometry{top=0.75in,left=0.75in,right=0.75in,bottom=0.75in} such as ChatGPT \cite{chatGPT}. In this paper, we conduct an empirical study by ``chatting'' with ChatGPT and discuss our findings in constructing a driving scenario domain ontology. Our contributions are as follows: 
\begin{itemize}[leftmargin=*]
    \item We are the first, to the best of our knowledge, to propose an empirical automation and semi-automation framework for domain knowledge distillation with LLMs.
    \item We discuss our key observations and recommendations covering the entire distillation lifecycle in depth. 
    \item We present our web-based domain ontology distillation assistant to facilitate runtime human supervision, addressing the key challenges faced in the automatic ontology distillation experiment.
\end{itemize}

This paper is organized as follows: Section \ref{sec:methodology} presents an overview of our empirical distillation framework, Section \ref{sec:experiment} demonstrates the application of the framework in the autonomous driving domain and discusses our key observations based on the distillation results, Section \ref{sec:ui} presents our web-based distillation assistant and Section \ref{sec:conclusion} concludes the paper.

%% file: Methodology.tex
\section{Distillation Framework Overview}\label{sec:methodology}

\subsection{Ontology 101}

The term \textit{Ontology} is defined as the description of domain \textbf{concepts} (often referred as \textbf{classes}, e.g., \textit{Car}, \textit{Lane}, \textit{Road}), the \textbf{properties} of the concepts (e.g., \textit{Color}, \textit{Size}, \textit{Length}), and the \textbf{relationships} (e.g., \textit{subclass\_of}, \textit{consists\_of}, \textit{drive\_on}) between the concepts. \cite{kendallOntologyEngineering}. The manual ontology construction process usually consists of the following steps (adapted from \cite{kendallOntologyEngineering,noy2001ontology}): 1) Define the application domain; 2) Define all the relevant classes; 3) Organize the classes in a superclass-subclass hierarchy; 4) Define the properties associated with each class; and 5) Define the relationships between each pair of classes.

In the next section, we present our empirical ontology distillation framework designed based on the required steps.

\subsection{Framework Overview}

\begin{figure}
    \centering
    \includegraphics[width=\columnwidth]{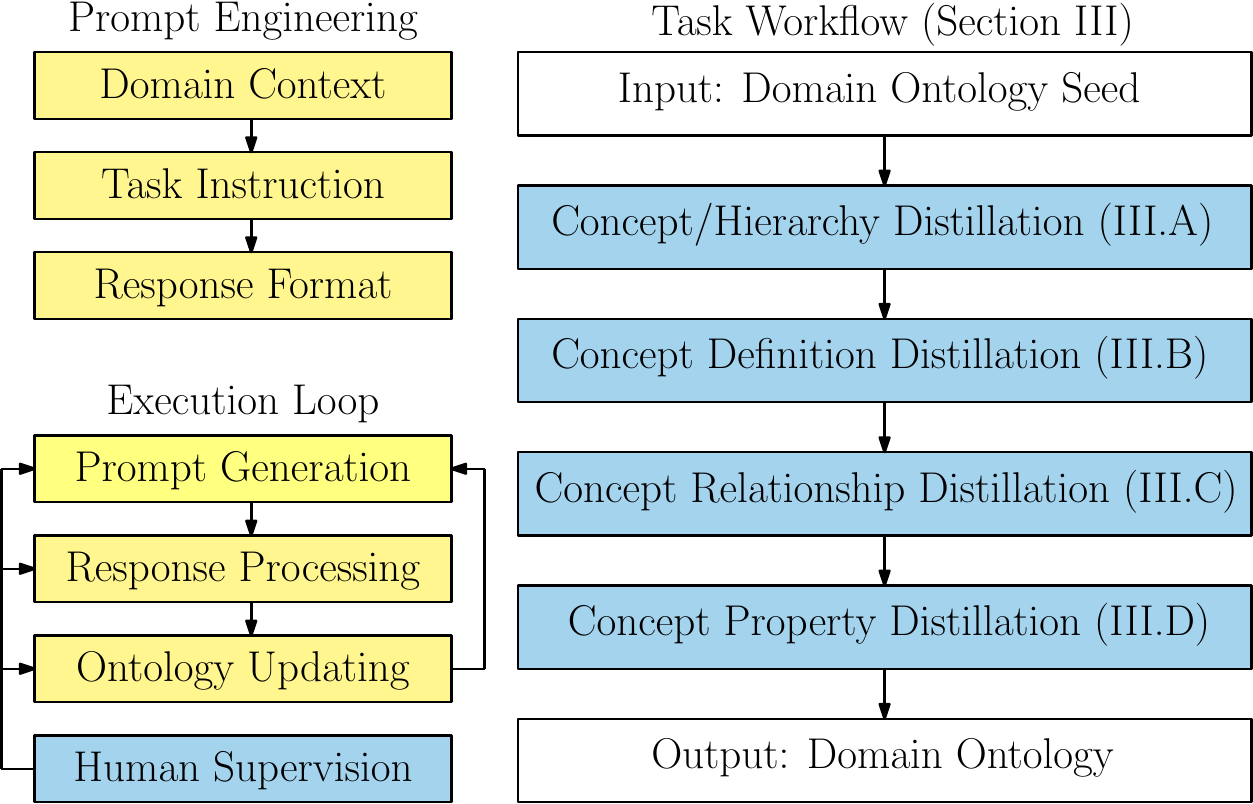}
    \caption{Our domain ontology distillation framework with three main components, i.e., Prompt Engineering, Task Workflow and Execution Loop.}
    \label{fig:automation_framework}
\end{figure}

The LLMs such as ChatGPT\cite{chatGPT} work in a question-answer (or instruction-response) mechanism, which enables us to extract and format the knowledge via ``prompt engineering''. To make this empirical study beneficial to most of the public, we limit the model to the browser version of ChatGPT \cite{chatGPT}, with default generation settings (e.g., temperature \cite{openai_APT_temperature}, and numbers of tokens \cite{openai_APT_tokens}) but provides unlimited free-tier usage. Figure~\ref{fig:automation_framework} presents our empirical distillation framework, consisting of three main components, i.e., Task Workflow, Prompt Engineering, and Execution Loop.

\textbf{Prompt Engineering} When ``chatting'' with ChatGPT, the prompt template consists of three parts, i.e., domain context (why), task instruction (what), and response format (how). The domain context part introduces the background context for the subsequent requests, e.g., ``\textit{I have a road driving scenario ontology as shown below ...}''. The task instruction part instructs the LLM on what information is expected, e.g., ``\textit{Add 10 new relevant concepts, terms or entities to the ontology ...}''. Lastly, to facilitate the automated processing of the responses, the response format part specifies the machine-readable format, e.g., ``\textit{Output the new ontology in DOT format.}''. Note that DOT \cite{DOT_wiki_page} (a graph description language, examples can be found in Figure~\ref{fig:concept_hierarchy_extraction_example}) is used in this study to describe the ontology class hierarchy as it is widely supported by major programming languages. 

\textbf{Task Workflow} We start with a seed ontology of the application domain and go through a list of distillation tasks (i.e., concept/hierarchy distillation, concept definition distillation, concept relationship distillation, and concept property distillation), wherein each task we repeatedly request new knowledge from ChatGPT to augment and improve the ontology. The reasons for such a workflow design are as follows: 

1) During our preliminary concept-distillation experiments, ChatGPT returns a wide range of concepts, including highly relevant, irrelevant, and sometimes duplicated ones, during the looped execution. As discussed in \cite{schafer2023adaptive, pearce2022examining}, with more specific context information and good examples come improved semantic accuracy and more focused responses. Thus, we need to provide illustrative examples in the prompt to distil those highly relevant concepts while eliminating the rest. This is essential, especially for the first request, as subsequent responses highly depend on the previous results, i.e., the butterfly effect applies. As a result, we introduce the seed ontology in the first request consisting of only highly abstract concepts but still sufficient to focus the scope and demonstrate the basic ontology structure, e.g., superclass-subclass relationship, in the DOT format.

2) The basis of an ontology is formed by the concepts and the classification hierarchy (organized by the pairwise superclass-subclass relationships between concepts). As we keep updating the ontology hierarchy, the location of individual concepts in the hierarchy also changes, and so do their definitions, non-hierarchical relationships (e.g., the \textit{drive on} relationship in \textit{vehicles drive on roads}) and properties. As a result, the distillation tasks for the hierarchy-dependent knowledge are performed after the ontology hierarchy has been constructed and fixed. 

\textbf{Execution Loop} In each task, there is an execution loop consisting of prompt generation, response processing and ontology updating, which continues until any stopping criterion is met, e.g., ChatGPT stops presenting new information or the ontology graph has reached a pre-defined breadth or depth. If ChatGPT returns irrelevant or erroneous results, the execution loop can be paused, repeated, reverted or resumed manually at any step to ensure satisfactory distillation results. In each step, we start a new conversation with ChatGPT instead of using the existing conversation sessions. Such a looped execution mechanism is proposed for the following reasons: 

1) It is impractical to extract all the information with one request due to the limit on the maximum number of tokens \cite{openai_APT_tokens} and occasional browser connection timeout exceptions (ChatGPT slowly generating a large body of text may encounter timeout error) per request. 

2) As ChatGPT memorizes its previous requests and responses in the same conversation session, it may return similar undesirable responses as in the previous responses. Hence, we design the prompt schema to be self-sufficient and start a new conversation for each request to avoid such scenarios. 

3) The looped execution mechanism improves the distillation quality and lessens the bufferfly effect by enabling manual supervision and early optimization. For example, in each step of the concept/hierarchy distillation task, instead of asking ChatGPT only to append new concepts while preserving the existing hierarchy, we request it to re-design the hierarchy from scratch considering all the concepts, explicitly requesting it to remove irrelevant concepts and merge duplicated ones. Such a step-wise re-design allows ChatGPT to optimize the hierarchy globally.  

In the next section, we will cover the details of each element in the empirical distillation framework by demonstrating the application in the autonomous driving domain and discuss our key observations and challenges.

%% file: Experiment.tex
\section{Domain Application of the Framework}\label{sec:experiment}

As mentioned, we apply the framework to construct an ontology in the road traffic domain for scenario-based V\&V of autonomous vehicles. Although domain experts (including our team) have already designed such an ontology manually as part of the OpenXOntology framework, we still need similar ontologies for many other transportation domains. This section presents our findings for the ontology distillation lifecycle based on our team's experiences in the OpenXOntology projects.

\subsection{Concept/Hierarchy Distillation} \label{sec:concept_hierarchy_extraction}

\begin{figure}
    \centering
    \includegraphics[width=\columnwidth]{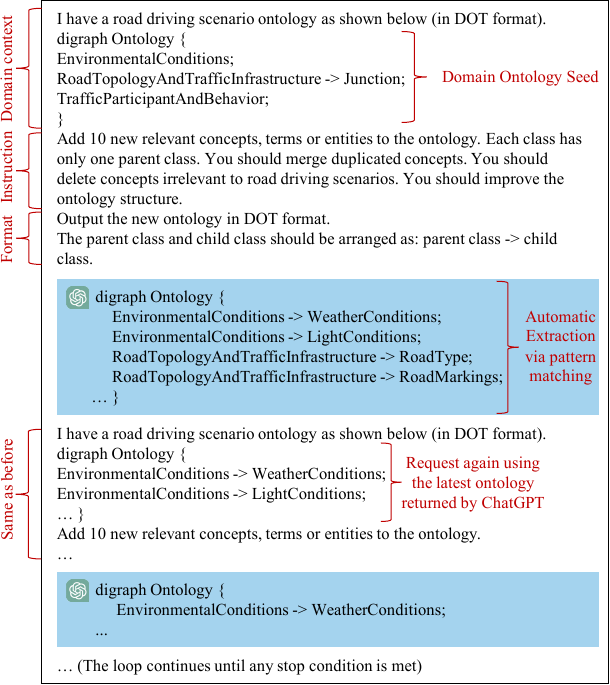}
    \caption{Concept/hierarchy distillation task chat example.}
    \label{fig:concept_hierarchy_extraction_example}
\end{figure}

\begin{figure}
    \centering
    \includegraphics[width=0.8\columnwidth]{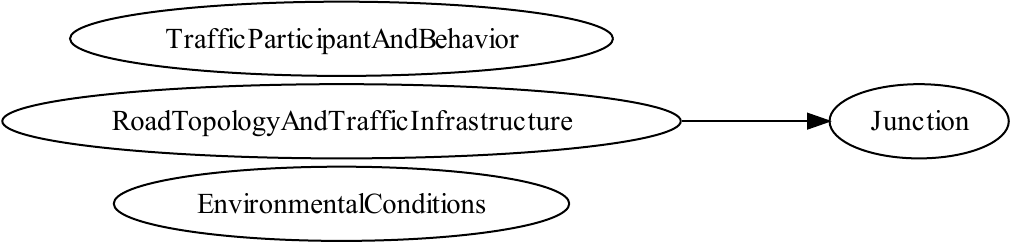}
    \caption{The seed ontology used in the autonomous driving domain application.}
    \label{fig:seed_ontology}
\end{figure}

\begin{figure}
    \centering
    \includegraphics[width=0.8\columnwidth]{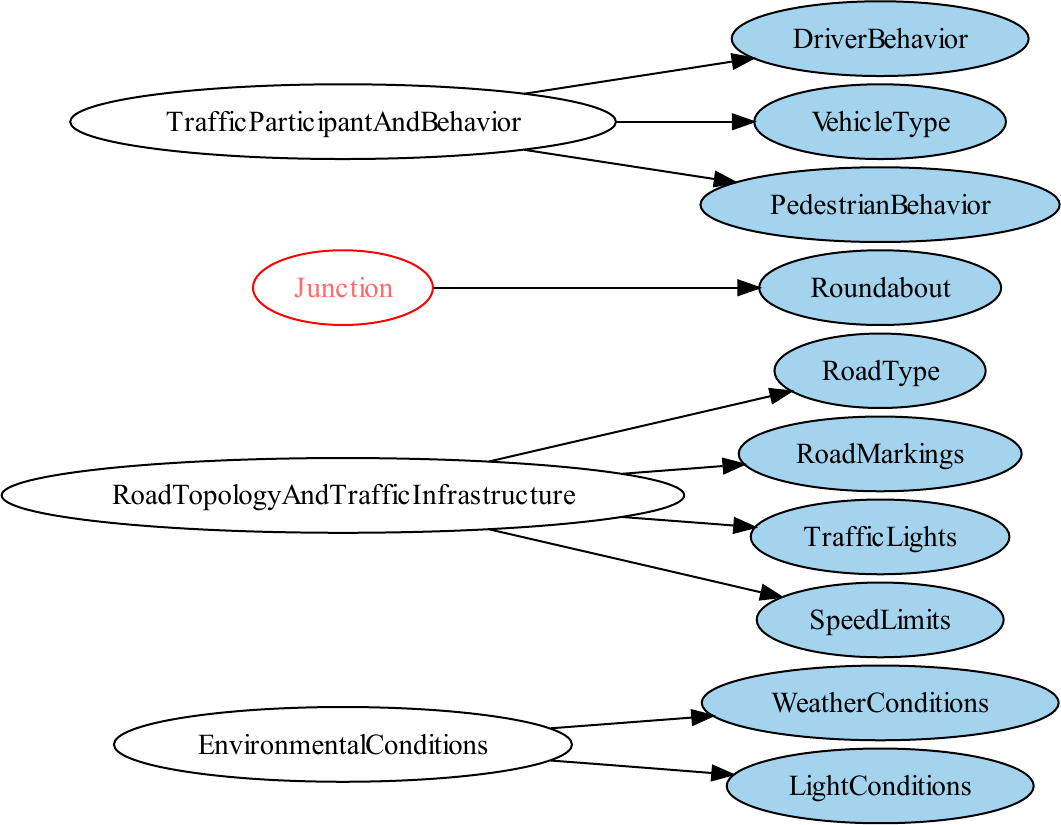}
    \caption{Concept/hierarchy task result after the first iteration. New concepts compared to the seed ontology are highlighted in blue.}
    \label{fig:hierarhcy_result_first_iteration}
\end{figure}

\begin{figure}
    \centering
    \includegraphics[width=\columnwidth]{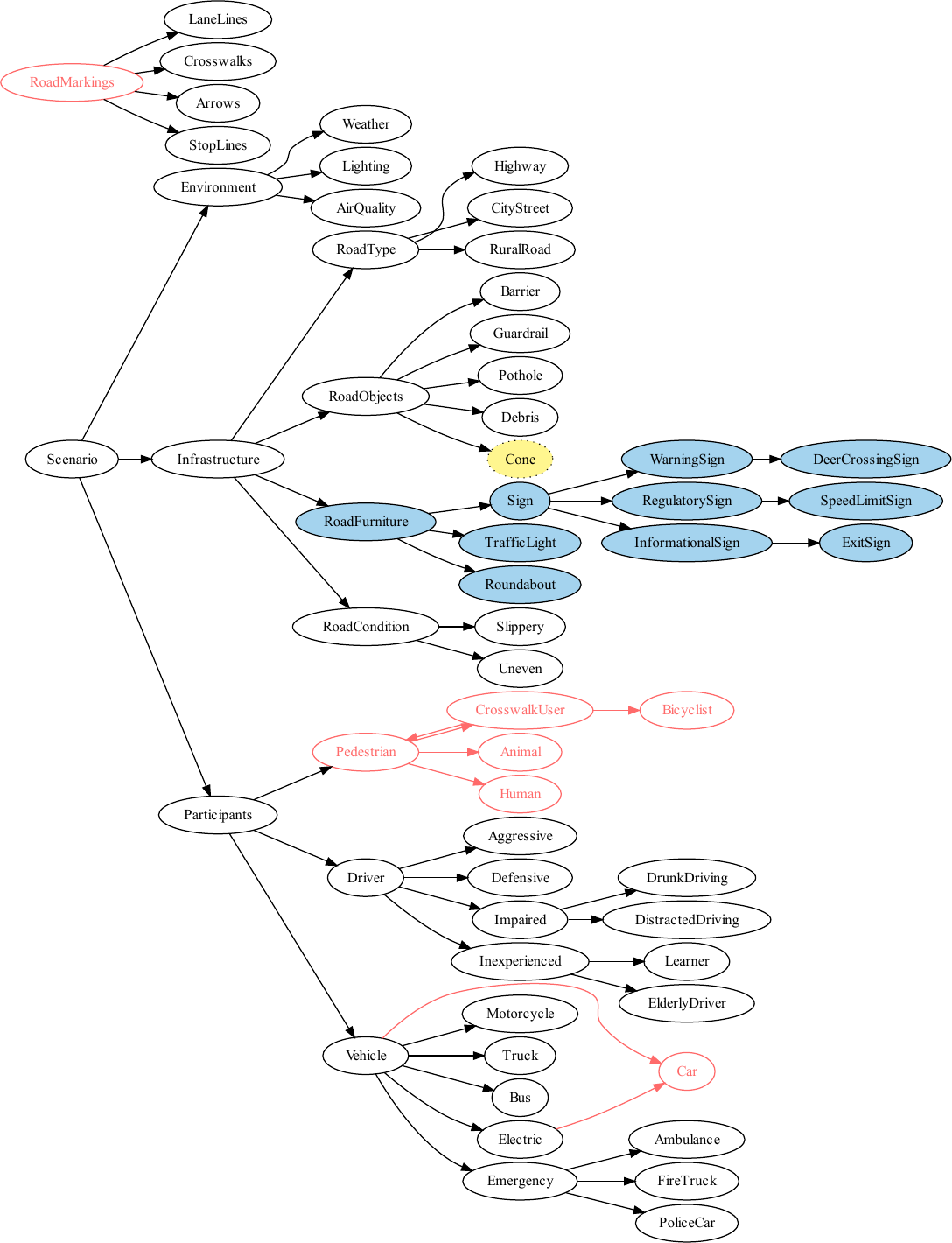}
    \caption{Concept/hierarchy task result after 10th iteration. New concepts compared to the 9th iteration are highlighted in blue.}
    \label{fig:hierarchy_result_tenth_iteration}
\end{figure}

Figure~\ref{fig:concept_hierarchy_extraction_example} presents a looped execution example of a concept/hierarchy distillation task. We design the seed ontology (Figure~\ref{fig:seed_ontology}) to include the three highly abstract seed concepts taken from OpenXOntology \cite{ASAM_OpenXOntology}, i.e., \textit{EnvironmentalCondition}, \textit{RoadTopologyAndTrafficInfrastructure} and \textit{TrafficParticipantAndBehavior}. In addition, the concept \textit{Junction} (a subclass of \textit{RoadTopologyAndTrafficInfrastructure}) is added intentionally as a superclass-subclass example in DOT format. 
During the looped execution, we use the same prompt template, only updating the ontology description part with the latest refined ontology by ChatGPT automatically extracted from the previous response. Due to limited space, we only present the distillation results after the first (Figure~\ref{fig:hierarhcy_result_first_iteration}) and the tenth (Figure~\ref{fig:hierarchy_result_tenth_iteration}) iteration.

In the first iteration, ChatGPT correctly introduces 10 new concepts to the seed ontology in the first iteration result, i.e., ``Driver Behavior', ``Vehicle Type'', ``Pedestrian Behavior'', etc. However, it gets confused by the concept name ``Road Topology And Traffic Infrastructure'' and generates four concepts, e.g., ``Road Type'' and ``Road Markings'', only related to ``roads''. As a result, it separates the ``Junction'' concept from its original category. This result is still considered semantically valid as the definitions of the seed concepts, including the ``Road Topology'' concept, are absent for ChatGPT.

In the tenth response, we have distilled many new concepts and a remarkable ontology hierarchy compared to the seed ontology. Based on the concept/hierarchy distillation process, we have the following observations:
\begin{tcolorbox}
\textbf{Observation 1:} ChatGPT may delete highly relevant concepts during iteration.
\end{tcolorbox} 
The equation $C_{N} = C_{0} + 10 \times N$, where $C_{N}$ is the total number of concepts distilled after $N$ iteration(s), does not necessarily hold as ChatGPT is specifically allowed to merge duplicate concepts and remove irrelevant concepts. Note that it removed the \textit{Cone} concept from the ninth iteration, which we believe should not have done so as \textit{Cones} are valid road objects. In addition, the seed concept \textit{Junction} has been removed from the entire ontology in one of the middle iterations when it becomes ``less relevant'' (as ChatGPT believes) to the ontology of that iteration.

\begin{tcolorbox}
\textbf{Observation 2}: ChatGPT tends to return highly cohesive concepts in each response.
\end{tcolorbox}
For example, in the tenth iteration, all the newly distilled concepts belong to the ``Road Furniture'' category. This behaviour is highly beneficial if, in the later stage, we would like to fine-tune a specific part of the ontology graph, for example, by asking ``\textit{Add 10 new relevant concepts under the Car category}''.

\begin{tcolorbox}
\textbf{Observation 3}: ChatGPT starts to overlook the details in the prompt as the prompt gets longer.
\end{tcolorbox}
With a bigger ontology graph comes longer request prompts according to our prompt engineering design, as we need to include the full ontology DOT description. As the prompt gets longer, ChatGPT starts to ignore the specific requirements. For example, we explicitly require that each concept has only one parent class (Figure~\ref{fig:concept_hierarchy_extraction_example}); However, ChatGPT disobeys by setting two parents (i.e., \textit{Vehicle} and \textit{Electric}) for the \textit{Car} concept (Figure~\ref{fig:hierarchy_result_tenth_iteration}). This might also be the reason for the undesirable removal of highly relevant concepts, e.g., \textit{Junction}.
\begin{tcolorbox}
\textbf{Observation 4}: It is impractical, if not impossible, to specify all the requirements during prompt engineering.
\end{tcolorbox} 
First, prompt engineering is a closed-loop process where prompts are improved iteratively based on the previous responses. Due to the randomness in the response, the prompt engineering process is also random. For example, in trial $T_1$, one may need to put one constraint $C_1$ to fix a response issue $Bug_1$; while in the rest of the trials $T_N$, one may never encounter $Bug_1$ although $C_1$ is absent in prompts. Imagine that one has collected a considerable number ($N$) of bug-fixing constraints $\bigcup_{i=1}^N C_i$, ChatGPT will likely fail to obey all the constraints as discussed before.
Moreover, lengthy constraints would shadow the ontology description content part and thus potentially result in ChatGPT overlooking some of the existing concepts or hierarchical relationships of the ontology. On the other hand, if one limits the number of constraints in the prompt, ChatGPT will inevitably return undesirable responses. For example, we do not specify that ontology hierarchy should be acyclic. In the tenth response, a loop in the ontology graph is formed between the concepts \textit{Pedestrian} and \textit{Crosswalk User}. Although the DOT language permits loops in directed graphs, it is sometimes confusing and undesirable for domain ontologies.
\begin{tcolorbox}
\textbf{Observation 5}: ChatGPT extends the ontology in a balanced (depth vs breadth) but a random way where the hierarchy grows in both breadth and depth.
\end{tcolorbox}
Our repeated experiments show that while the ontologies of different experiment trials share many concepts, e.g., \textit{car}, \textit{Pedestrian}, and \textit{Driver}, etc., those concepts have different hierarchical locations and relative distillation orders across the trails. Hierarchies across many trails are also different. This suggests we may need to repeat the automated distillation experiment until we achieve a preferable concept set and hierarchical structure. 

With the above observations, we recommend a manual examination and error fixing during or at least immediately after the \textit{Concept/Hierarchy Distillation} task. For example, the ontology after the tenth loop is manually fixed, before the subsequent \textit{Concept Definition Distillation} task, with the following minor modifications: a) set concept \textit{Road Markings} a sub-class of concept \textit{Infrastructure}; b) remove the concept \textit{Electric}; c) remove the concept \textit{Crosswalk User} and set \textit{Bicyclist} a sub-class of \textit{Pedestrian}.

\subsection{Concept Definition Distillation} \label{sec:concept_definition_extraction}

\begin{figure}
    \centering
    \includegraphics[width=\columnwidth]{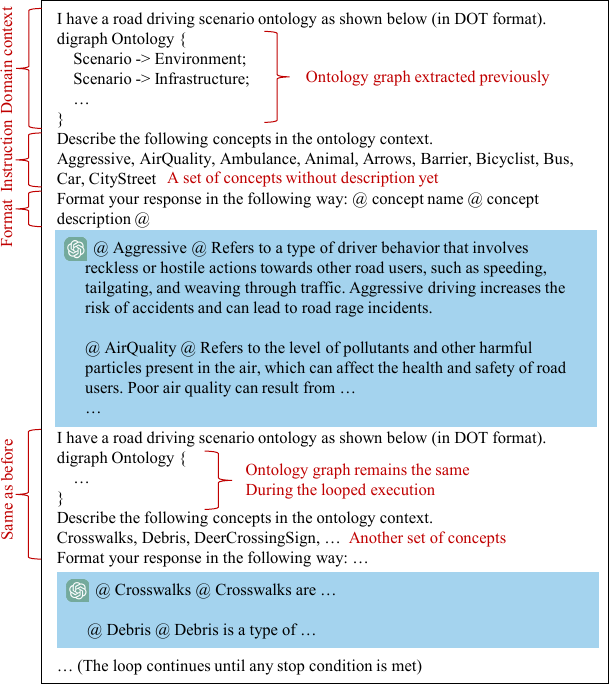}
    \caption{Concept definition distillation task chat example}
    \label{fig:concept_definition_extraction_example}
\end{figure}

\begin{figure}[t]
    \centering
    \includegraphics[width=\columnwidth]{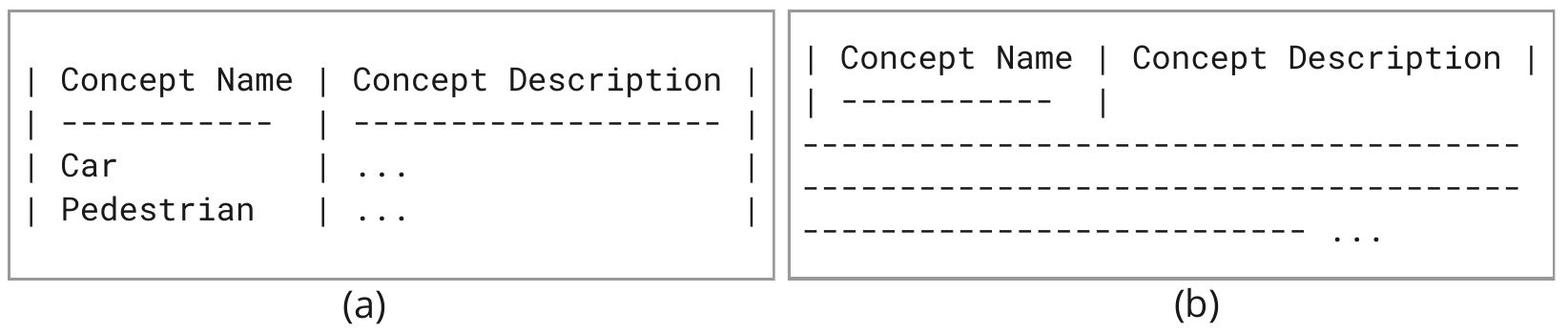}
    \caption{ChatGPT table formatting bug: (a) normal table format output. (b) incorrect table output with abundant ``- - -'' causing timeout or render errors}
    \label{fig:table_format_error}
\end{figure}

Figure~\ref{fig:concept_definition_extraction_example} shows a concept definition distillation example. We first present the previously distilled ontology hierarchy and ask ChatGPT to define a fixed number of concepts such that it can take the entire ontology hierarchy into consideration. This is the primary reason why we distil definitions after hierarchy. We request ten concept definitions per conversation during the experiment to avoid issues such as timeout. By default, ChatGPT prefers to format its output into a markdown table (Figure~\ref{fig:table_format_error} (a)). However, a markdown formatting bug is often encountered during the loop execution phase, where ChatGPT keeps printing the ``- - -'' symbols without stopping. To address this issue and eliminate the number of tokens used for formatting purposes (i.e., tokens are wasted by the ``- - -'' characters), we propose to use comma-separated values (CSV) with the separator ``@'' as it is unlikely to appear in the concept names and definitions. The distillation loop stops until all the concepts are defined. There are 56 concepts in the revised ontology hierarchy; hence, it stops after six loop iterations.  

As we initiate a new conversation (instead of continuing with the previous conversation session), ChatGPT gives the concept definition in different styles. Selected responses of different styles are listed below:

\begin{itemize}[leftmargin=*]
    \item \textit{Bicyclist}: \textit{Refers to a person who is riding a bicycle on or near the road.} \inlinegray{(definition)} \textit{Bicyclists have the same rights and responsibilities as other road users and must follow the same traffic laws.} \inlinegray{(additional information)}

    \item \textit{Driver}: \textit{Driver is part of the Participants category in the road driving scenario ontology.} \inlinegray{(superclass reference)} \textit{A driver is a person who operates a vehicle on the road.} \inlinegray{(definition)} \textit{Drivers can be categorized by their driving behaviour or experience, such as aggressive, defensive, impaired, or inexperienced.} \inlinegray{(subclass reference)}

    \item \textit{Aggressive}: \textit{Refers to a type of driver behaviour that involves reckless or hostile actions towards other road users, such as speeding, tailgating, and weaving through traffic.} \inlinegray{(definition with examples)} \textit{Aggressive driving increases the risk of accidents and can lead to road rage incidents.} \inlinegray{(additional information)}

\end{itemize}

To facilitate discussion, we label (in grey) all the sentences based on their semantic nature. We have the following observations:

\begin{tcolorbox}
\textbf{Observation 6}: ChatGPT generally defines each concept with a random mixture of key components, i.e., definition, additional information, reference to the concept's superclasses, and reference to the sub-classes.
\end{tcolorbox}
The combination styles remain coherent within the same conversation session and may vary across different conversation sessions. Our further experiments show that the combination style can be customized easily with prompt engineering, e.g., \textit{``in the concept description, describe its definition, its relative position in the ontology hierarchy, and provide any additional relevant information''}.    

\begin{tcolorbox}
\textbf{Observation 7}: The definitions with illustrative examples may signal further concept/hierarchy distillation.
\end{tcolorbox}
In many concept definitions similar to the \textit{Aggressive} concept, ChatGPT also illustrates the concept with concrete examples, while those examples are not present in the ontology yet. This indicates that ChatGPT has additional knowledge regarding the concepts, and further concept/hierarchy distillation can be conducted.

\subsection{Concept Relationship Distillation} \label{sec:concept_relationship_extraction}

\begin{figure}
    \centering
    \includegraphics[width=\columnwidth]{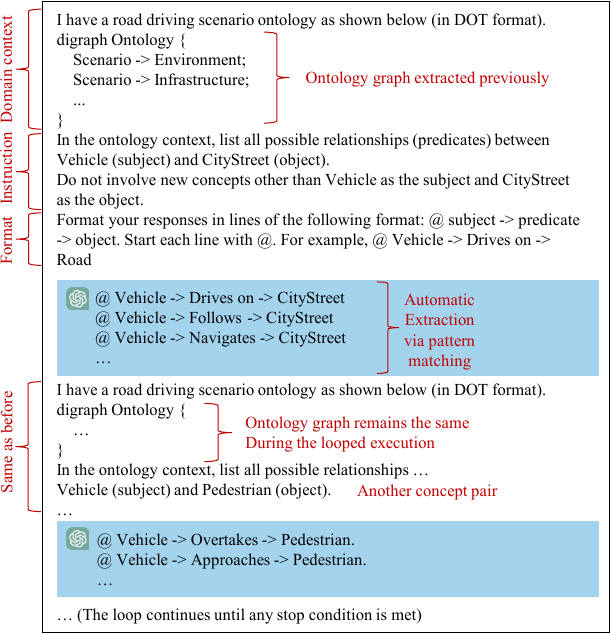}
    \caption{Concept relationship distillation task chat example}
    \label{fig:concept_relationship_extraction_example}
\end{figure}

This section discusses the results of the non-hierarchical relationship (relationships other than superclass-subclass relationship) distillation task. 
We aim to distil two main types of relationships, i.e., the \textbf{inter-concept relationship} and \textbf{intra-concept relationship}. The inter-concept relationships lie between different concepts, e.g., \textit{Vehicle} and \textit{Road}, while intra-concept relationships exist between the same concepts, e.g., \textit{Vehicle} and \textit{Vehicle}. In this study, we limit the scope to pairwise relationships by explicitly specifying only two concepts in the following form: $Subject~Concept \rightarrow Relationship~Predicate \rightarrow Object~Concept$, 
e.g., \textit{Vehicle} $\rightarrow$ \textit{Drives on} $\rightarrow$ \textit{Road}. 

Figure~\ref{fig:concept_relationship_extraction_example} shows an execution example of the relationship distillation task. We adjusted the format request to avoid erroneous responses like: ``\textit{@ Emergency @ Uses @ Ambulance @ to respond to incidents affected by poor @ AirQuality @.}'' when ``@'' is used as the delimiter. Selected relationship responses are listed in Table~\ref{tab:relationship_extraction_result}. Each relationship distillation result is a union of five independent executions. We have the following observations:

\begin{table*}[]
\caption{Distilled relationship examples for both intra (left) and inter (right)-concept relationships}
\label{tab:relationship_extraction_result}
\resizebox{\textwidth}{!}{%
\begin{tabular}{ll|ll}
\hline
\multicolumn{1}{c}{\textbf{\begin{tabular}[c]{@{}c@{}}Subject/\\ Object\end{tabular}}} &
  \multicolumn{1}{c|}{\textbf{Intra-Concept Relationship}} &
  \multicolumn{1}{c}{\textbf{\begin{tabular}[c]{@{}c@{}}Subject/\\ Object\end{tabular}}} &
  \multicolumn{1}{c}{\textbf{Inter-Concept Relationship}} \\ \hline
\begin{tabular}[c]{@{}l@{}}Environment/\\ Environment\end{tabular} &
  \begin{tabular}[c]{@{}l@{}}Affects,  Influences,  Determines,  Modifies,  Depends on, Impacts,  Changes,  Interacts with,  \\ Alters\end{tabular} &
  \begin{tabular}[c]{@{}l@{}}Environment/\\ Infrastructure\end{tabular} &
  \begin{tabular}[c]{@{}l@{}}Affects, Alters, Changes, Conditions, \\ Determines, Impacts, Influences, Interacts \\ with, Modifies, Shapes\end{tabular} \\ \hline
\begin{tabular}[c]{@{}l@{}}Vehicle/\\ Vehicle\\\\(highlighted \\are new \\compared\\ to Car/Car)\end{tabular} &
  \begin{tabular}[c]{@{}l@{}}Accelerates past, Blocks, \textbf{Brakes suddenly in front of}, \textbf{Changes lane}, Changes lane behind, \\ Changes lane in front of, \textbf{Changes lane to}, \textbf{Collides into}, Collides with, Competes with, \\ \textbf{Cooperates with}, \textbf{Crosses path with}, Cuts off, Decelerates behind, Drafts behind, \\ \textbf{Drives alongside}, \textbf{Drives in front of}, Drives next to, Follows, \textbf{Gives way to}, Honks at, \\ Overtakes, \textbf{Parks behind}, \textbf{Parks in front of}, Parks next to, Passes, Pulls over for, Races, \\ Races with, \textbf{Shares the road with}, Signals to, Stops behind, Swerves to avoid, Tailgates, \\ \textbf{Turns left in front of}, \textbf{Turns right in front of}, \textbf{Yields to}\end{tabular} &
  \begin{tabular}[c]{@{}l@{}}Vehicle/\\ TrafficLight\end{tabular} &
  \begin{tabular}[c]{@{}l@{}}Approaches, Follows signal, Follows the \\ signal of, Halts before, Ignores, Ignores \\ the signal of, Obeys, Observes, Passes, \\ Proceeds on green, Proceeds through, Runs, \\ Stops at, Waits at, Waits at red, Waits for\end{tabular} \\ \hline
\begin{tabular}[c]{@{}l@{}}Car/\\ Car\\\\(highlighted \\are new \\compared\\ to Vehicle/\\Vehicle)\end{tabular} &
  \begin{tabular}[c]{@{}l@{}}Accelerates past, \textbf{Avoids}, \textbf{Avoids collision with}, Blocks, \textbf{Causes traffic jam with}, \textbf{Changes} \\\textbf{lane behind}, \textbf{Changes lane in front of}, Changes lanes behind, Changes lanes in front of, \\ Collides with, \textbf{Comes into view of}, Competes with, \textbf{Crashes into}, \textbf{Creates gap for}, Cuts off, \\ Decelerates behind, Drafts behind, \textbf{Drives beside}, Drives next to, \textbf{Drives past}, \textbf{Enters} \\\textbf{intersection with}, \textbf{Exits intersection with}, Follows, \textbf{Follows too closely behind}, \textbf{Follows too}\\ \textbf{closely to}, \textbf{Gets cut off by}, \textbf{Gets passed by}, \textbf{Gets stuck behind}, Honks at, \textbf{Lets in}, \\ \textbf{Merges behind}, \textbf{Merges in front of}, \textbf{Navigates around}, Overtakes, Parks next to, Passes, \\ \textbf{Passes by}, \textbf{Passes on the left/right of}, Pulls over for, Races, \textbf{Races against}, Races with, \\ \textbf{Rear-ends}, Signals to, \textbf{Signals to turn behind}, \textbf{Signals to turn in front of}, Stops behind, \\ \textbf{Stops next to}, Swerves to avoid, Tailgates, \textbf{Tows}\end{tabular} &
  \begin{tabular}[c]{@{}l@{}}FireTruck/\\ TrafficLight\end{tabular} &
  {\color[HTML]{333333} \begin{tabular}[c]{@{}l@{}}Activates, Affects, Approaches, Changes \\ direction at, Damages, Ignores, Obeys, \\ Passes, Proceeds after stopping at, Stops at, \\ \textbf{Turns on}\end{tabular}} \\ \hline
\begin{tabular}[c]{@{}l@{}}Aggressive/\\ Aggressive\end{tabular} &
  {\color[HTML]{333333} \begin{tabular}[c]{@{}l@{}}"There are no possible relationships between Aggressive and Aggressive because Aggressive \\ is not defined as a concept with any subtypes or attributes ..."\end{tabular}} &
  \begin{tabular}[c]{@{}l@{}}TrafficLight/\\ Vehicle\end{tabular} &
  {\color[HTML]{333333} \begin{tabular}[c]{@{}l@{}}Affects, Controls, Determines, Dictates, \\ Dictates actions of, Directs, Governs, Guides, \\ Indicates, Influences, Interacts with, Modifies, \\ Modifies behavior of, Regulates, Signals\end{tabular}} \\ \hline
\end{tabular}%
}
\end{table*}

\begin{tcolorbox}
\textbf{Observation 8}: The relationship distillations on any pairs of concepts, regardless of the concepts' relative hierarchical positions, are equally important. 
\end{tcolorbox}
Given a distilled ontology hierarchy of $N$ concepts, there are $N^2$ unique ordered subject-object concept pairs. We may extract the relationship of every concept pair when $N$ is trivial. However, as $N$ gets bigger, such a complete iteration can become expensive. Ideally, one may consider distilling relationships of concepts at greater heights (if we consider the ontology hierarchy as a tree) only as the relationships of the superclass shall be the union of all the sub-classes relationships, e.g., $Vehicle \rightarrow Vehicle$ include many distilled relationships with $Car \rightarrow Car$ (Table~\ref{tab:relationship_extraction_result}). However, in practice, this is only partially true for the following reasons: 

1) The relationship distillation results are random and independent in different conversation sessions regardless of the concept being a super or sub-class in the ontology context. The results for the superclass may be a part, union, intersection, or mix of the results for the sub-classes. For example, the \textit{Car} $\rightarrow$ \textit{Tows} $\rightarrow$ \textit{Car} relationship may never, although it should, be distilled for the Vehicle $\rightarrow$ \textit{Vehicle} pair. 

2) High-level super-class pairs, e.g., \textit{Environment} $\rightarrow$ \textit{Environment} and \textit{Environment} $\rightarrow$ \textit{Infrastructure} tend to induce abstract relationships, such as \textit{Affects}, \textit{Influences}, and \textit{Modifies} etc. While the abstract relationships are valid, they may be less useful in applications. For example, when we design testing scenarios, the term \textit{Influences} can be too broad as we need to specify how to \textit{influence} in concrete scenarios. 

3) Concrete sub-class pairs can potentially distil specialized relationships, e.g., \textit{FireTruck} $\rightarrow$ \textit{Turns on} $\rightarrow$ \textit{TrafficLight} is highly unlikely to appear for \textit{Vehicle} $\rightarrow$ \textit{TrafficLight} pair. 

\begin{tcolorbox}
\textbf{Observation 9}: The intra-concept relationship and inter-concept relationship distillation are equally important.
\end{tcolorbox}
Depending on the nature of the concept, ChatGPT may fail to return any intra-concept relationships (e.g., \textit{Aggressive} $\rightarrow$ \textit{Aggressive} as it regards \textit{Aggressive} as ``a property dependent on other concepts''. However, during inter-concept relationship distillation, ChatGPT can return relationships such as \textit{Aggressive} $\rightarrow$ \textit{Causes} $\rightarrow$ \textit{Emergency}, \textit{Aggressive} $\rightarrow$ \textit{Negatively impacts response time of} $\rightarrow$ \textit{Emergency} and 
\textit{Aggressive} $\rightarrow$ \textit{Interferes with the ability of} $\rightarrow$ \textit{Emergency} \textit{(to reach their destination quickly and safely)}.

\begin{tcolorbox}
\textbf{Observation 10}: Post-processing of the responses is often needed.
\end{tcolorbox}
To keep the ontology concise and organized, we note that post-processing is often necessary, for example, to merge synonyms (e.g., \textit{Races} vs \textit{Races With} vs \textit{Races Against}, \textit{Affects} vs \textit{Influences} vs \textit{Changes}, \textit{Tailgates} vs \textit{Follows too closely behind}), active-passive pairs (e.g., \textit{Passes} vs \textit{Gets passed by}), define relationship groups (e.g., \textit{Parks behind} vs \textit{Parks in front of} vs \textit{Parks next to}, \textit{Turn left in front of} vs \textit{Turn right in front of}) and filter unnecessary relationships (e.g., \textit{Shares the road with}).

\subsection{Concept Property Distillation}
The property distillation task shares many common characteristics with previous tasks, e.g., the property inheritance between the superclass and subclass. Due to limited space, similar observations are not discussed in this section.

%% file: UI.tex
\section{Web User Interface}\label{sec:ui}
\begin{figure*}
    \centering
    \includegraphics[width=\textwidth]{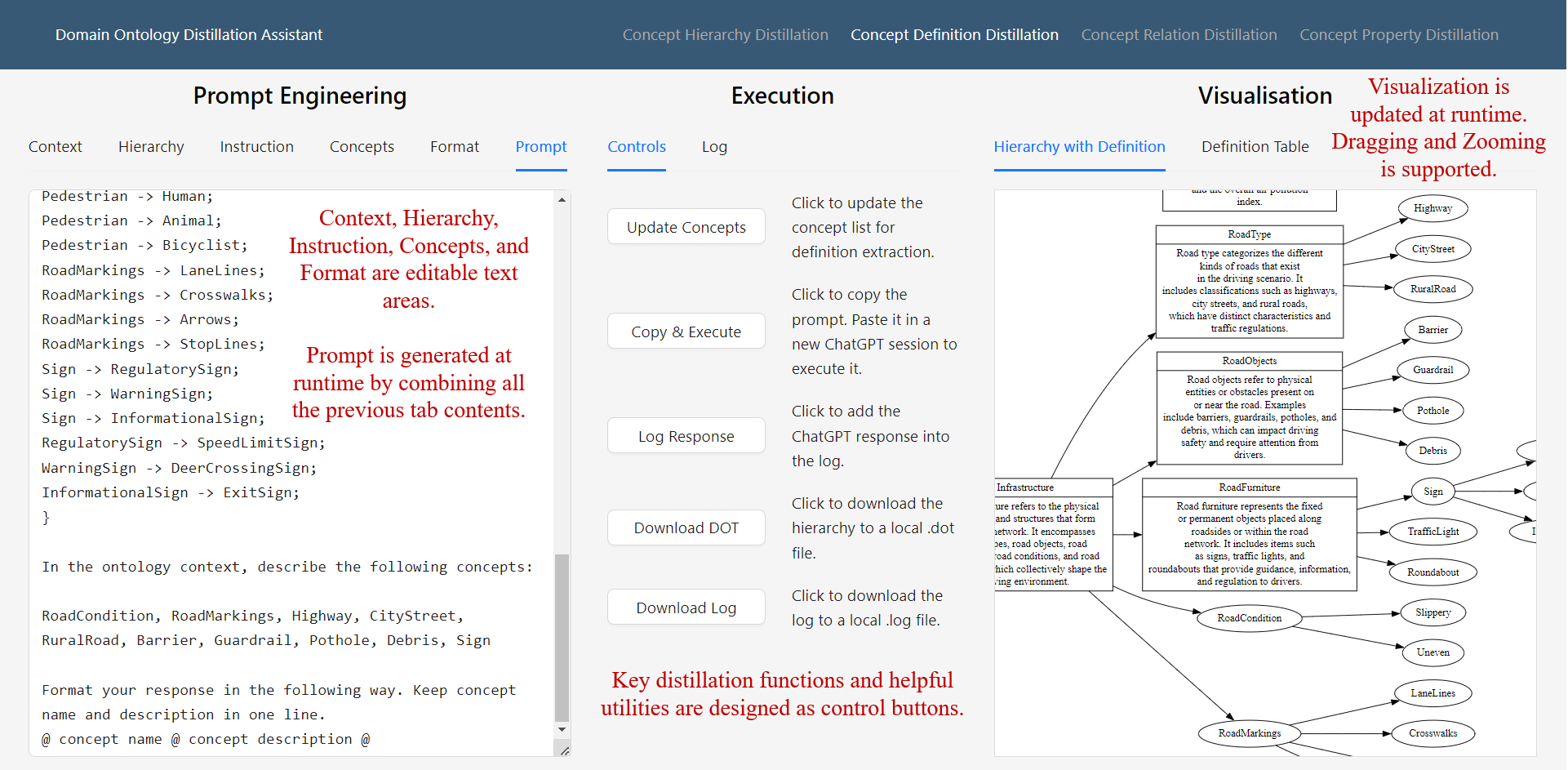}
    \caption{Website user interface (the concept definition distillation page) of the Domain Ontology Distillation Assistant}
    \label{fig:website-ui}
\end{figure*}

Based on our empirical study results and observations, a fully automated ontology distillation process is possible. However, it may lead to unpredictable and irrelevant ontology results due to the randomness in the responses and the butterfly effect. Manual supervision and early intervention are still required to guarantee distillation quality, improve efficiency and save potential costs (e.g., from repeated trials). To facilitate this, we develop a web-based domain ontology distillation assistant as shown in Figure~\ref{fig:website-ui}. The website has four sub-pages corresponding to the four distillation tasks. In the prompt engineering section, all the essential components are rendered as independent editable text areas for maximum flexibility, e.g., the user may change the instruction part from ``\textit{Add 10 new relevant concepts, ..., to the ontology}'' to ``\textit{Add 10 new concepts under the Vehicle class}''. The execution log contains the complete history of both prompts and ChatGPT's responses in each iteration. After ChatGPT's response is logged, the entire log is parsed, the ontology is updated, the visualization is refreshed, and the prompt for the next iteration will be generated. To facilitate manual supervision and early intervention, the user can then decide whether to continue the next step or make necessary adjustments to the ontology or prompt during the entire execution loop. Currently, extensive engineering effort is underway to improve the assistant tool's usability and design across transportation application domains, and we are pleased to open-source it soon.

%% file: Conclusion.tex
\section{Conclusion}\label{sec:conclusion}

This paper presents our empirical domain knowledge distillation framework using ChatGPT and discusses our observations from the framework application experiments in the autonomous driving domain. The key finding is that: 1) with proper design of prompt engineering and execution flow, fully automated domain knowledge (in the ontology format) distillation is possible. However, due to the randomness in the response and the butterfly effect, the quality of fully automated distillation results is not guaranteed. To address this, we develop a web-based assistant to enable manual supervision and early intervention at runtime. We hope our findings and tools inspire future research toward revolutionizing the engineering processes of knowledge-based systems across domains.